# Spatial-Temporal Block and LSTM Network for Pedestrian Trajectories Prediction

Dan Xiong[1]    Fan Li[1]    Yipeng Nie[2]

[1]State Key Laboratory of Advanced Design Manufacturing for Vehicle Body, Hunan University, Changsha 410082, China.

[2]Sany Automobile Manufacturing Co., LTD, Sany industry town, Economic and Technological Development Zone, Changsha 410100, China

xiongdan891@163.com, lifandudu@163.com, nieyp@sany.com.cn

## Abstract

*Pedestrian trajectory prediction is a critical to avoid autonomous driving collision. But this prediction is a challenging problem due to social forces and cluttered scenes. Such human-human and human-space interactions lead to many socially plausible trajectories. In this paper, we propose a novel LSTM-based algorithm. We tackle the problem by considering the static scene and pedestrian which combine the Graph Convolutional Networks and Temporal Convolutional Networks to extract features from pedestrians. Each pedestrian in the scene is regarded as a node, and we can obtain the relationship between each node and its neighborhoods by graph embedding. It is LSTM that encode the relationship so that our model predicts nodes trajectories in crowd scenarios simultaneously. To effectively predict multiple possible future trajectories, we further introduce Spatio-Temporal Convolutional Block to make the network flexible. Experimental results on two public datasets, i.e. ETH and UCY, demonstrate the effectiveness of our proposed ST-Block and we achieve state-of-the-art approaches in human trajectory prediction.*

## 1 Introduction

In recent years, autonomous mobile platforms such as self-driving cars and social service robots have developed rapidly. It is significant to predict the pedestrian trajectory in order to prevent the damage caused by the collision of mobile platform to people. For example, the correct prediction of pedestrian trajectory can help the real-time decision monitoring system to issue early warning and take preventive measures rapidly. The world we live in is inherently structured. It is comprised of components that interact with each other in space and time, leading to a spatio-temporal composition. The next direction of the pedestrian trajectory is related to many factors, such as avoiding collisions, observing traffic regulations, adjusting social etiquette, etc. Macroscopically, the pedestrian and the social force between two nodes are regarded as a node and an edge as shown in Figure 1. The number of nodes and edges in a scene which change over time is dominate the position of each node in future.

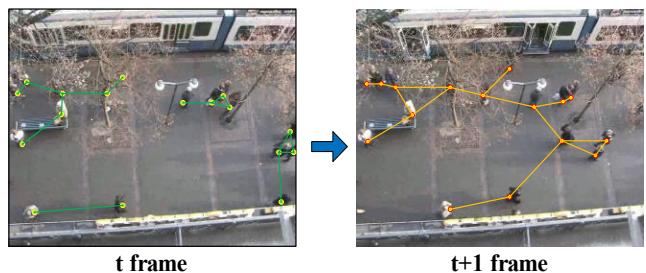

Figure 1: At t frame and t+1 frame, the distribution of position and the social forces between nodes of pedestrians. Each person in the scene can be regarded as a node. Edges have different influence on two nodes connected at different times.

Method in forecasting pedestrian trajectory generally fall into two categories in previous work. The first approach is to divide pedestrian trajectories into many clusters. Researchers utilized gaussian process regression (GPR) to cluster pedestrian trajectories[2,3,4,5]. However, when the pedestrian intention is ambiguous, the output of the regression algorithm tend to be the average of the different intention trajectories which is obviously not adapted to the real scene. The second method is to estimate the next step of pedestrian trajectory according to the pedestrian trajectory history. Kalman filter is the most commonly used method to track and predict the next state in a linear system[1,6,7]. Whether the former or the latter, uncertainty in pedestrian decisions leads to difficulties in prediction. The prediction of pedestrian trajectory should be probabilistic rather than deterministic. Generative Adversarial Networks (GANs) have been recently developed can overcome the difficulties in approximating intractable probabilistic computation and behavioral inference[18].

Following the success of neural networks has boosted research on pattern recognition and data mining. A host of researches such as object detection[28] and speech recognition[27] utilize convolutional neural networks (CNNs), recurrent neural networks (RNNs), and autoencoders[29] to extract informative feature sets instead of heavily relying on handcrafted feature engineering. They learn a target node's representation by propagating neighbor information in an iterative manner until a stable fixed point is reached. With the development of Recurrent Neural Network (RNN) models for sequence prediction tasks, many researchers try to leverage data-driven method based on long-short term memory networks (LSTM)[8,10,11,12,13,14,17,18] to learn social behavior for increase robustness and accuracy in multi-target tracking problems.

Our model influenced by the recent success of Spatio-Temporal Graph[26,32,33,36,37] and also Graph Embedding[39] in different real-world problems, we combine the spatial position in each pedestrian with the surroundings and the previous temporal pedestrian trajectory to predict all pedestrian trajectories in crowd scenarios simultaneously. ST-LSTM focuses on the adjustment for current LSTM states, which is quite different from existing RNN-based approaches. To adaptively extract social effects from neighborhoods and make the network flexible, we will introduce the Spatio-Temporal Convolutional Block which can stack or extend the graph features of the network according to the scale and complexity.

Our main contributions are as follows:

(1) The approach we present in this paper can be viewed as a data driven approach by combining the spatial position in pedestrians and pedestrians, pedestrian and the surrounding and the previous temporal pedestrian trajectory.

(2) We present a spatio-temporal convolutional block that explicitly captures the global interaction of all the pedestrians in the scene and the local interaction with the static objects, and we propose a new graph embedding for each pedestrian trajectory with LSTM networks. Our model can achieve better prediction than state-of-the-art methods.

## 2 Related Works

The single-person trajectory is related to many factors, such as the nature and spatial distribution of the surrounding obstacles. It is forecasting human behavior that can be grouped as learning to predict human-space interactions or human-human interactions. In this section, we give a brief review of related work.

Since the moving trajectories of human are nonlinear and complex, it is hard to be fully expressed by tradition method like hand-craft rules. The existing approaches follow different strategies to solve this problem. [30] tracked individual trajectories using contextual information without the need to learn the structure of the scene. [22] combined contextual information with information about the posture and body movement of the pedestrian to improve the classifications results. [9] found possible paths to probabilities from a set of possible goals, a map of the scenario and the initial position of the pedestrian. Such predictions can't be satisfied with scene interactions between the human and the scene or other humans. In contrast, our methods are designed for more natural scenarios where goals are open-ended, unknown or time-varying and where human interact with each other while dynamically predicting in anticipation of future trajectories.

### 2.1. LSTM networks for sequence prediction

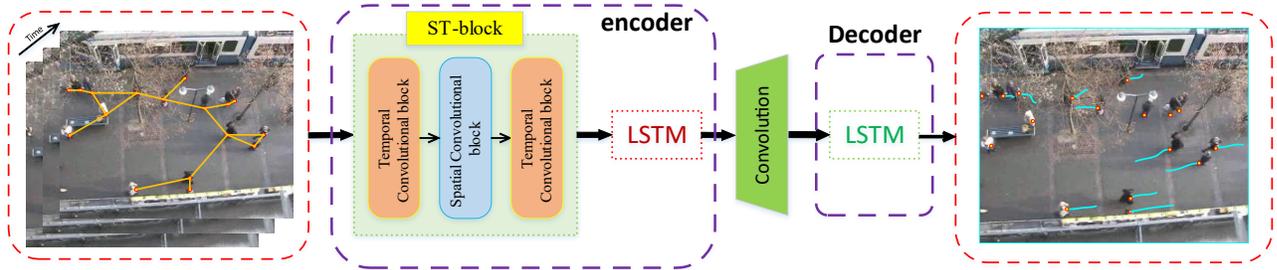

Figure2: The model we both use two temporal convolutional networks and one spatial convolutional network for a ST-Block to extract features following by LSTM nets. The nodes of the graph represent the pedestrians, and the edges capture their spatio-temporal interactions.

The success of LSTM networks in modeling non-linear temporal dependencies in sequence learning and generation tasks inspired many researchers. Intuitively, the trajectories of pedestrians can be considered as time sequence data, so LSTMs can be used for predicting trajectories of pedestrians.

A novel pooling layer is presented by [17], where the hidden states of neighboring pedestrians are shared together to jointly reason multiple people. State-Refinement LSTM (SR-LSTM)[12] extends [17] with new pooling mechanisms and activated the utilization of the current intention of neighbors to improve the prediction precision. [8] proposed a bidirectional LSTM architecture to yield multiple prediction trajectories with different probabilities towards different destination regions in the scene. [10] proved the adapted LSTM model capability of capturing and representing a variety of complex paths is great for generating data, but somewhat obstructive for prediction tasks, such as endpoint prediction. [13] used a multi-layer perceptron to map the location of each pedestrian to a high dimensional feature space for measuring the spatial affinity between two pedestrians. Nonetheless, these models lack predictive capacity as they do not take into account scene context. Later, hierarchical LSTM-based network is proposed by [14] to consider both the influence of social neighborhood and scene layouts to fully model social scale human-human interactions. [31] leveraged the path history of all the agents in a scene and the scene context information, using images of the scene. However, these methods fall short as instead of treating pedestrian's future movements as a distribution of locations, they only predict a single path.

## 2.2. Encoder-Decoder framework

Encoder-decoder learns the state of a person and stores their history of motion and generates target sequences. Naïve recurrent models could fail in modeling interactions between pedestrians correctly. [20] introduce an RNN Encoder-Decoder framework which uses variational autoencoder (VAE) for trajectory prediction. [11] extended the classical encoder-decoder framework in sequence to sequence modelling to incorporate both soft attention as well as hard-wired attention. A sequence-to-sequence LSTM encoder-decoder is trained [15,18,26], which encodes observations into LSTM and then decodes as predictions.

## 2.3. Spatial Temporal Graph Convolutional Networks

Although the deep RNN architectures are remarkably capable at modeling sequences, lack an intuitive high-level spatio-temporal. So far, spatio-temporal graph has been used for prediction in many fields. Researchers applied the spatio-temporal graph to accurately model traffic prediction[32,37] and model dynamic skeletons[35,36]. In addition, [24] introduced Diffusion Convolutional Recurrent Neural Network (DCRNN) which a deep learning framework that incorporates both spatial and temporal dependency in the traffic flow. [26] proposed an approach which combined the power of high-level spatio-temporal graphs and sequence learning success of Recurrent Neural Networks (RNNs) to model human motion and object interactions. However, few works have been trying to combine both spatial-temporal and LSTM networks to model pedestrian sequences and predict their trajectories from distribution of pedestrian and time steps. Our model is able to take into account other surrounding pedestrians and is able to generate multiple plausible paths using a GAN module. Two key difference between our model ST-LSTM and Social-STGCNN is that features we extracted link each other by ST-Blocks and draw interactions by LSTM networks to directly model pedestrian trajectories.

## 2.4. Graph Embedding

Graph is an important data representation which appears in a wide diversity of real-world scenarios. Effective graph analytics provides a deeper understanding of the pedestrian datasets. Graph embedding converts graph data into a low dimensional, compact, and continuous feature space. The key idea is to preserve the topological structure, vertex content, and other side information. For example, the second-order proximity in a graph can be preserved in the embedded space by maximizing the probability of observing the neighborhood of a node conditioned on its embedding[39]. Social-STGCNNs extract both spatial and temporal information from the graph creating a suitable embedding[33]. But our work draws on the experience of a novel graph embedding framework[38] which encodes the topological structure and node content for spatial features in a scene to predict the trajectories of pedestrians.

The model we both use two temporal convolutional networks and one spatial convolutional network for a ST-Block to extract features following by LSTM nets. The nodes of the graph represent the pedestrians, and the edges capture their spatio-temporal interactions. Our method moves beyond the limitations of previous methods by automatically learning both the spatial and temporal patterns from graph.

## 3 Method

We regard the scene per frame as a whole and the pedestrian walking space as a connected graph. We extract the spatial and temporal features of each scene by Graph Convolutional Networks and Temporal Convolutional Networks respectively, so that it can form a spatio-temporal block. Because of LSTM network success in sequence modeling, we use LSTM networks to encode spatio-temporal features of difference states which in order to ensure temporal dependency of sequences. The LSTMs are interconnected in a way that the resulting architecture captures the structure and interactions of the st-graph. In order to jointly reason across multiple people to share information across LSTMs, our work add the convolutional module. After the decoding the information, our work can achieve the purpose of pedestrian trajectory detection accurately.

### 3.1. Problem Definition

Our work is to jointly reason and predict the future trajectories of all the pedestrians simultaneously involved in a scene. We assume that there are $N$ pedestrians $p_1, …, p_N$ at $T$ time frame in a scene. The spatial location (absolute coordinate) of the $i^{th}$ pedestrian $p_i(i \in [1, N])$ at time t is denoted as $P_t^i = (x_t^i, y_t^i)$, where $x_t^i \in [1, X]$, $x_t^i \in [1, Y]$, and X, Y is the spatial resolution of video frames. The input trajectory is defined as $P_{1 \to T_{obs}}$ from time steps $t = 1, …, T_{obs}$ and the future trajectory (ground truth) can be defined similarly as $P_{T_{obs} \to T_{pred}}$ from time steps $t = T_{obs}, …, T_{pred}$.

### 3.2. Spatio-Temporal graph network

The Spatio-Temporal graph network consists two parts for extracting graph features. The first part is Graph Convolutional Networks (GCN) for extracting spatial features. GCN extract global neighborhood of each node in graph by defining convolution operators on graphs. The model iteratively aggregates the neighborhood embeddings of nodes and uses the embeddings and embedded functions obtained in the previous iteration to achieve new embeddings. Its superiority is that the aggregated embedding of the local neighborhood makes network scalable and allow embeddin-g a node to describe the global neighborhood through multiple iterations. The other part is Gated CNNs[25] for Extracting Temporal Features. we employ entire convolutional structures on time axis to capture temporal dynamic behaviors of pedestrian flows. This specific design allows parallel and controllable training procedures through multi-layer convolutional structures formed as hierarchical representations.

### 3.3. Graph Representation of Pedestrian Trajectories

We introduce the construction of the graph representation of pedestrian trajectories. We construct a set of spatial graphs $G$ representing the relative locations of pedestrians in a scene at each time step $t$, $G=\{V_t, E_t, X_t\}$ where $V_t$ is a set of nodes in the graph $G_t$, $V_t=\{v_t^i\}$, $i \in 1,...,n$, and n is the sum of pedestrians at the time step t. $E_t$ is a set of edges which indicate the connecting line between nodes, $E_t = \{e_t^{ij}\}$, $i,j \in 1,...,n$. $X_t$ is represent the content features that depicts the attributes of nodes themselves associated with each node $v_t^i$. The topological structure of graph $G_t$ can be represented by an adjacency matrix $A$. The weighted adjacency matrix $A$ is a representation of the graph edges attributes. The kernel function maps attributes at $v_t^i$ to a value $A_{ij}$ attached to $e_t^{ij}$, where $A_{ij}=1$ if $e_t^{ij} \in E_t$, otherwise $A_{ij}=0$.

### 3.4. Graph Convolutional Networks for extracting spatial features

We are inspired by ARGA[38] which leveraged GCN[40] to encode node structural information and node feature information at the same time. We employ the absolute coordinates of each pedestrian in a graph as a node for the sake of encode the topological structure and node content in a graph to a compact representation, on which a decoder is trained to reconstruct the graph structure.

Our graph convolutional network (GCN) extends the operation of convolution to graph data in the spectral domain, and learns a layer-wise transformation by a spectral convolution function:

$$Z^{l+1}=f(Z^l, A|W^l) \quad (1)$$

where $Z^l$ is the input for convolution, and $Z^{l+1}$ is the output after convolution. $Z^0=X \in R^{n \times m}$, $n, m$ represents the aggregate of nodes and features respectively. $W^l$ is a matrix of filter parameters we need to learn in the neural networks. Each layer of our graph convolutional network expressed with the function:

$$Gconv_0(Z^l, A|W^l) = ReLU(\widetilde{D}^{-\frac{1}{2}} \widetilde{A} \widetilde{D}^{-\frac{1}{2}} Z^l W^l) \quad (2)$$

where $\widetilde{A}=A+I$ and $\widetilde{D}_{ii}=\Sigma_j \widetilde{A}_{ij}$. $I$ is the identity matrix of $A$. $ReLU$ is activation function. The graph spatial embedding is constructed with a two-layer GCN:

$$\mathbf{H} = Gconv_1(Gconv_0(Z, A|W^0), A|W^1) \quad (3)$$

where **H** denotes the network embedding matrix of a graph, and the first layer $Gconv_0$ and the second layer $Gconv_1$ use $ReLU$ and linear activation function respectively.

### 3.5. Temporal Convolutional Networks for extracting temporal features

In order to extract the temporal features of the scene at each t moment accurately, we select TCN as the starting point of the deep network[41]. TCN take a stacked sequential data as input and predict a sequence as a whole. the TCN is achieved by two measures, it used a 1D fully-convolutional network (FCN) architecture which each hidden layer is the same length as the input layer, and zero padding of length is added to keep subsequent layers the same length as previous ones and used causal convolutions where an output at time $t$ is convolved only with elements from time $t$ and earlier in the previous layer. To put it simply:

$$TCN = 1D\ FCN + causal\ convolutions \quad (4)$$

Temporal model can focus more on the salient regions of the scene and the more relevant neighboring pedestrians in order to predict the future state truly.

### 3.6. Spatio-temporal Convolutional Block

We are inspired by[37] that the fusion of spatial and temporal features can better handle the time series of graph structures, and the spatio-temporal convolutional block can stack or extend the graph features of the network according to the scale and complexity to make the network flexible. The spatial layer in the middle is to bridge two temporal layers as shown in ST-Block part of Fig 2, which can propagate from graph convolution to fast spatial state through time convolution for scale compression and feature squeezing. The input and output of ST-Blocks are three-dimensional tensor, which contain batch size, max nodes and sequence length. In addition, layer normalization is utilized within every ST-Block to prevent overfitting.

| Datasets | Baselines | | | | | | | Our model |
|---|---|---|---|---|---|---|---|---|
| | Linear [17] | LSTM [17] | S-LSTM [17] | S-GAN [18] | SoPhie [31] | SR-LSTM [12] | Social-STGCNN [33] | |
| ETH | 1.33/2.94 | 1.09/2.41 | 1.09/2.35 | 0.81/1.52 | 0.71/1.43 | 0.63/1.25 | 0.64/1.11 | **0.57/1.02** |
| HOTEL | 0.39/0.72 | 0.86/1.91 | 0.79/1.76 | 0.72/1.61 | 0.76/1.67 | **0.37**/0.74 | 0.49/0.85 | 0.41/**0.73** |
| UNIV | 0.82/1.59 | 0.61/1.31 | 0.67/1.40 | 0.60/1.26 | 0.54/1.24 | 0.51/1.10 | **0.44**/0.79 | **0.38**/0.84 |
| ZARA1 | 0.62/1.21 | 0.41/0.88 | 0.47/1.00 | 0.34/0.69 | **0.30**/0.63 | 0.41/0.90 | 0.34/0.53 | 0.31/**0.47** |
| ZARA2 | 0.77/1.48 | 0.52/1.11 | 0.56/1.17 | 0.42/0.84 | 0.38/0.78 | 0.32/0.70 | **0.30/0.48** | 0.33/0.51 |
| AVG | 0.79/1.59 | 0.70/1.52 | 0.72/1.54 | 0.58/1.18 | 0.54/1.15 | 0.45/0.94 | 0.44/0.75 | **0.40/0.72** |

Table 1: Quantitative results of several methods compared to ST-LSTM model are shown. Error metrics reported are ADE / FDE in meters. All the methods predict trajectories for 12 frames using 8 frames' observed trajectories. Our model consistently outperformed the baselines, due to the combination of ST-graph and LSTMs in a model setting.

### 3.7. LSTM based Encoder-Decoder framework

We need a compact representation which combines information from different encoders to effectively reason about social interactions. Long Short-Term Memory (LSTM) networks have been proved successful in sequence modeling. In a basic LSTM network architecture, given an input sequence represented by $(x_1,...,x_T)$, the output sequence $y_t$ can be obtained by iteratively computing Eqs. (1) and (2) for $t=1,...,T$:

$$h_t = LSTM(h_{t-1}, x_t, W) \quad (5)$$
$$y_t = W_{hy} h_t + b_y \quad (6)$$

where the $W$ terms denote the different weight matrices, $b_y$ denotes the bias vector for the output $y_t$, and $h$ denotes the hidden state.

### 3.8. Encoder module

By following previous works, we employ LSTM networks to encode the ST-block information for each pedestrian and capture the dependency among the ST-blocks of different states. Encoder learns the spatial and temporal states of pedestrians and stores their history of trajectories. These embeddings are used as input to the LSTM cell of the encoder at time t introducing the following recurrence:

$$b_{st}^t = \{S^t, T^t\} \quad (7)$$
$$h_b^t = LSTM(h_b^{t-1}, b_{st}^t, W_{encoder}) \quad (8)$$

Where $b_{st}^t$ is an ST-block, $S^t, T^t$ is spatial features block and temporal features block respectively. $h_b^t$ is hidden state at t and the LSTM weights ($W_{encoder}$) are shared between all people in a scene.

### 3.9. Decoder module

Recent work 33 uses the bivariate Gaussian distribution as positive sample of pedestrian position. Using this method to determine that positive samples can be applied to common population densities. Besides, we use the memory characteristics of LSTM network to decoder spatial and temporal features of all pedestrians and reconstruct the graph data to predict nodes locations.

$$h_{dec}^t = LSTM(\gamma(P_i, h_{dec}^{t-1}), b_{st}^{t-1}, W_{decoder}) \quad (9)$$
$$(\hat{x}_i^t, \hat{y}_i^t) = \gamma(h_{dec}^t) \quad (8)$$

Where the LSTM weights are denoted by $W_{decoder}$ and $\gamma$ is an MLP with *ReLU* non-linearity. We apply L2 loss on the predicted trajectory which measures how far the predicted samples are from the actual ground truth. We also use graph information both the graph structure A and content information X to train a prediction layer to determine whether there is an interaction between two nodes.

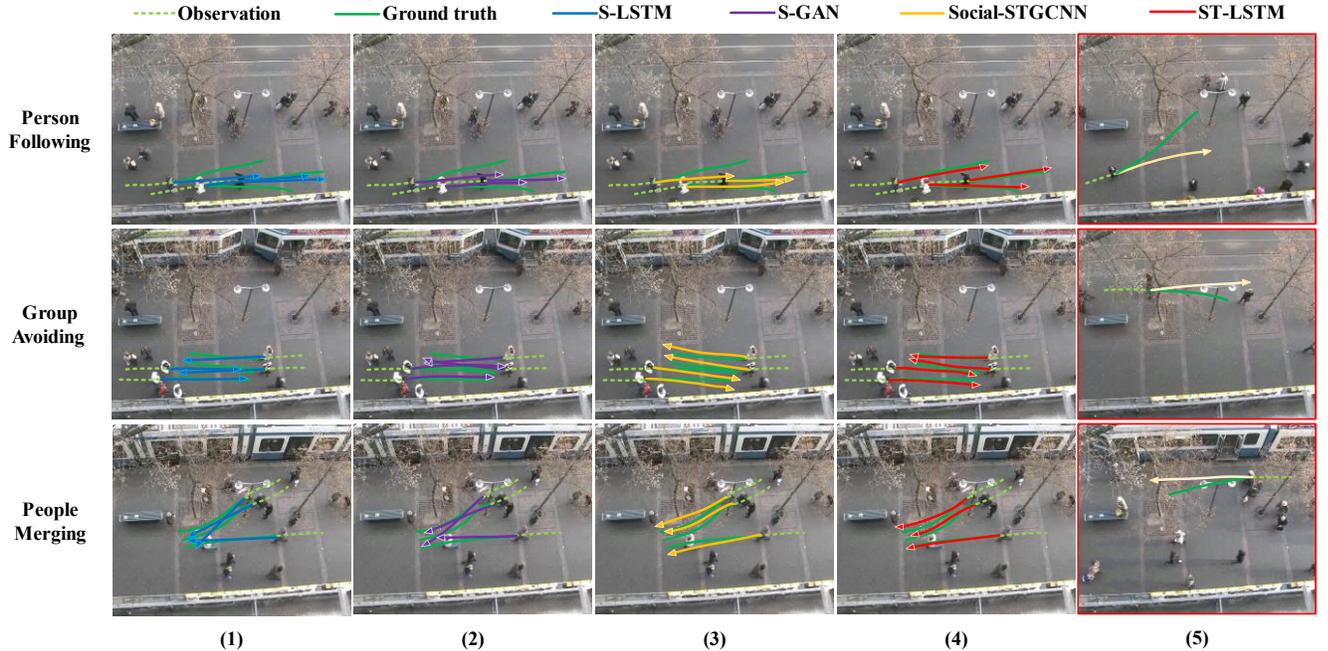

Figure 6: Examples of diverse predictions from three baselines and our model. Each row shows three sets include person following, group avoiding and people merging of observed trajectories; Columns show four different models for each scenario which demonstrate different types of socially acceptable behavior. Our model predicts globally consistent and socially acceptable trajectories for all people by data-driven manner in the scene. Our prediction trajectory is closer to the real ground and avoids collisions by changing their direction. The last column is failure predictions.

| Datasets | Ground Truth | Linear | S-GAN | SoPhie | Our model |
|---|---|---|---|---|---|
| **ETH** | 0.000 | 3.137 | 2.509 | 1.757 | 1.258 |
| **HOTEL** | 0.092 | 1.568 | 1.752 | 1.936 | 1.237 |
| **UNIV** | 0.124 | 1.242 | 0.559 | 0.621 | 0.581 |
| **ZARA1** | 0.000 | 3.776 | 1.749 | 1.027 | 0.871 |
| **ZARA2** | 0.732 | 3.631 | 2.020 | 1.464 | 0.983 |
| **AVG** | 0.189 | 2.670 | 1.717 | 1.361 | 0.986 |

Table 2: Average % of colliding pedestrians per frame for each of the scenes in BIWI/ETH. A collision is detected if the euclidean distance between two pedestrians is less than 0.10m.

$$p(\widehat{A}_{ij} \mid Z) = \prod_{i=1}^{n} \prod_{j=1}^{n} p(\widehat{A}_{ij} \mid z_i, z_j) \quad (10)$$

$$p(\widehat{A}_{ij}=1 \mid z_i, z_j) = sigmoid(z_i^T, z_j) \quad (11)$$

We denote the decoder loss function as $\mathcal{L}_0$ which minimize the construction error of graph data:

$$\mathcal{L}_0 = E_{q(Z\mid(X,A))} \log p(\widehat{A} \mid Z)] \quad (12)$$

## 4 Experiments

**Datasets.** We evaluate our model on two publicly available datasets: ETH[19] and UCY[21]. ETH contains two scenes named ETH and HOTEL, while UCY contains three scenes named ZARA1, ZARA2 and UNIV. Both datasets involve 4 different scenes which consists of 1536 pedestrians in crowded settings. The trajectories in datasets are sampled every 0.4 seconds. These datasets consist of real world human trajectories with rich human-human interaction scenarios such as group behavior, people crossing each other, collision avoidance and groups forming and dispersing.

**Evaluation Metrics and baselines.** Similar to prior work[1]

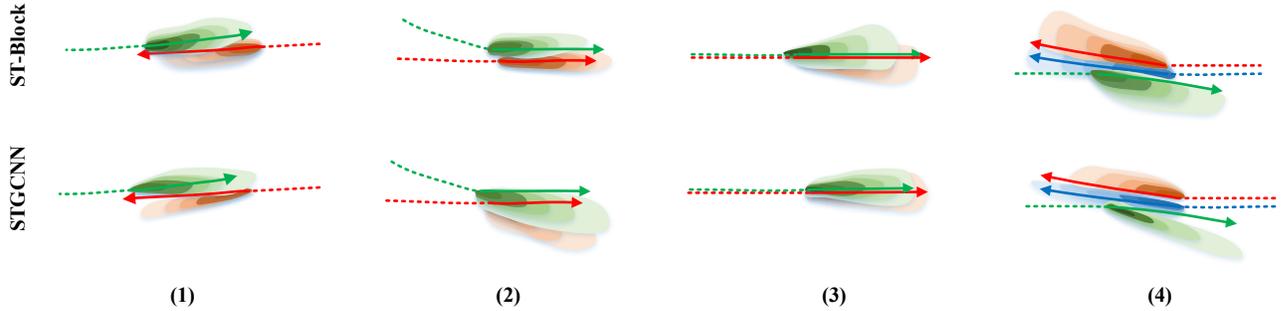

Figure 5: Comparison between our model and SGAN-P in four collision avoidance scenarios: two people meeting (1), one person meeting a group (2), one person behind another (3), and two people meeting at an angle (4). For each example we draw 300 samples from the model and visualize their density and mean.

[18]we use two error metrics:

1. Average Displacement Error (ADE): Average L2 distance between ground truth and our prediction over all predicted time steps.

$$ADE = \frac{\sum_{n \in N} \sum_{t \in T_p} \|\hat{p}_t^n - p_t^n\|^2}{N \times T_P}, t = T_{pred} \qquad (13)$$

2. Final Displacement Error (FDE): The distance between the predicted final destination and the true final destination at end of the prediction period $T_{pred}$.

$$FDE = \frac{\sum_{n \in N} \|\hat{p}_t^n - p_t^n\|^2}{N}, t = T_{pred} \qquad (14)$$

**Baselines**. We compare against the following baselines:

1. Linear: A linear regressor that estimates linear parameters by minimizing the least square error.

2. LSTM: A simple LSTM with no pooling mechanism.

3. S-LSTM: Each person is modeled via an LSTM with the hidden states being pooled at each time step using the social pooling layer.

4. S-GAN: Predictive models that applies generative modeling to S-LSTMs.

5 SR-LSTM: The States Refinement (SR) module for LSTM networks.

6. SoPhie: Model which combine path history and the scene context information by framework based on Generative Adversarial Network (GAN)

7. Social-STGCNN: The model is composed of a series of GCN layers followed by TXP-CNN layers.

**Model configuration and training setup.** We Transform the input sequence into a three-dimensional (batch size, sequence length, max nodes) graph vector and use LSTM as the RNN for both decoder and encoder. Our model contained a ST-Block which comprised of one GCN layer and two TCN layers. The output from ST-Block fed into an Encoder with a hidden dimension of 32 and a Decoder with a hidden dimension of 64. The final layer is passed through a convolutional layer. We set a training batch size of 128 and the model was trained for 250 epochs using Stochastic Gradient Descent with an initial learning rate of 0.001 to optimize model. The interval of trajectory sequences is set to 0.4 seconds. We take 8 ground truth positions as observation, and predict the trajectories of following 12 times steps, which follows the setting of[18]. We use PReLU 23 as the activation function $\sigma$ across our model which allows a slightly different span of the negative hidden states along training batches. this has proved its benefit for the model prediction performance.

### 4.1 Quantitative Evaluation

We follow similar evaluation methodology as [18]. We compare our method with the prior methods on two metrics ADE and FDE in Table 1. The evaluation task is defined to be performed over 8 seconds, using the past 8 positions consisting of the first 3.2 seconds as input, and predicting the remaining 12 future positions of the last 4.8 seconds.

Four datasets are evaluated by performing a leave-one-out cross-validation policy. During the experiments on these datasets, we train our network using 4 subsets and test it on the remaining 1 subset.

As expected, linear model which cannot model the complex social interactions between different humans is only to model straight paths and does especially bad in case of longer predictions ($T_{pred}$ = 12). Both LSTM and S-LSTM perform much better than the linear baseline as they can model more complex trajectories, especially in S-LSTM. S-LSTM utilizes the previous neighboring LSTM states by social pooling. S-GAN provides an improvement to this LSTM baseline and acquire various trajectories by generative model. SoPhie use social context for better predictions, but it is not enough to truly understand the interactions in a scene. SR-LSTM refines the current LSTM states in order to timely capture changes of the others' intention and make suitable adjustment. Unfortunately, its performance is still not effective. The previous state of art on the ADE and FDE metric is Social-STGCNN[33] with an error of 0.44 and 0.75. Our model has an error of 0.40 on the ADE metric and 0.72 on the FDE which are about 9.1% and 4% less than the state of the art respectively. It can be observed that the minimum FDE is considerably lower than minimum FDE generated by S-GAN model, SR-LSTM and SoPhie, due to our model awareness of spatial and temporal features.

**Collision rate.** To better demonstrate the superiority of our model in avoiding pedestrian collision events, we evaluated the percentage of near-collisions from the predicted trajectory. We will treat it as a collision if two pedestrians get closer than the threshold of 0.10m. We have calculated the average percentage of pedestrian near collisions across all frames in each of the BIWI/ETH scenes. These results are presented in Table 3. We sampled first 8 seconds of each trajectory among 30 random agents from the test datasets. It is obviously that the performance of our model is much better than three baseline methods.

### 4.2 Qualitative Analysis

We further investigate the ability of our architecture to model how social interactions impact future trajectories. Fig. 4 demonstrates the affects that spatial and temporal features can have in correcting erroneous predictions. We visualize three different scenes include person following, group avoiding and people merging, comparing our model to the ground truth pedestrian movements. In order to show the accuracy of our model prediction, we choose the three most representative models to compare the predicted trajectories with my model. Here is a specific description of the prediction results in three scenarios.

**People following.** (Row 1) A common scenario is when a person is walking behind someone. Followers may continue to follow or speed up beyond the front. The trajectory prediction of this scene mainly considers the trajectory of the follower, after all, the leader can't perceive the trajectory behind it to avoid collision. Because of the ground truth is closest linear equation, all the trajectory predictions give moderate future predictions and relatively low errors. But our prediction is the least different from the actual.

**Group avoiding.** (Row 2) When two people or people walk opposite each other, avoiding collisions is a basic reaction. A person will plan his path ahead of time according to the distribution of his neighborhood. S-LSTM and S-GAN utilizes the previous neighboring LSTM states to predict, but their predictions were insensitive to neighborhoods and caused a slight collision. ST-GCNN and our model did show desirable prediction.

**People Merging.** (Row 3) Two people converge from different directions to a path, the main people will pay attention to the path on the branch line, and the people on the branch line will slow down the pace of convergence. There are obvious collisions on trajectories of S-LSTM and S-GAN. Social-GCNN lead to abnormal deviation from the true path even though avoid collision. Our model employs both LSTM effective memory and spatial features to effectively navigate our predictions match the real ground. However, some samples show undesired behaviors such as collision or divergence in the last column.

In order to understand that our model prediction is closer to real ground, we compare how our ST-Block and Social-GCNN perform in four common social interaction scenarios (see Figure 4). We refer to the method from SGAN[18] to create these scenarios to evaluate the models and we used models trained on real world data. For each setup, we draw 300 samples and plot an approximate distribution of trajectories along with average trajectory prediction. Plainly, our ST-Block adjusts the advancing direction to avoid possible collisions much better than Social-GCNN in four scenes.

## 5 Conclusion

Due to the high nonlinearity and complexity of pedestrian flow, traditional methods cannot satisfy the requirements of mid-and-long term prediction tasks and often neglect spatial and temporal dependencies. To take full advantage of spatial features, we use Spatio-Temporal Block(ST-Block) to capture adjacent relations among the graph with employing Long Short-Term Memory(LSTM) network. We show the efficacy of our method on several complicated real-life scenarios where social norms must be followed.